\documentclass[runningheads]{llncs}
\usepackage[T1]{fontenc}
\usepackage{graphicx}
\usepackage{amsmath}
\usepackage{algorithm}
\usepackage{amssymb}
\usepackage{algpseudocode}
\usepackage{orcidlink}
\usepackage{booktabs}
\usepackage{subcaption}
\usepackage{siunitx}
\usepackage{comment}
\usepackage{bm}           
\usepackage{tikz}
\usetikzlibrary{shapes.geometric, arrows.meta, positioning, calc}

\newcommand{\ACC}{\mathrm{ACC}}
\newcommand{\AUC}{\mathrm{AUC}}

\newcommand{\CLACC}{\mathrm{CLSACC}}
\newcommand{\NMI}{\mathrm{NMI}}
\newcommand{\MI}{\mathrm{MI}}            

\newcommand{\FSDEM}{\mathrm{FSDEM}}

\newcommand{\OT}{\mathrm{OT}}
\newcommand{\TP}{\mathrm{TP}}
\newcommand{\TN}{\mathrm{TN}}
\newcommand{\FP}{\mathrm{FP}}
\newcommand{\TPR}{\mathrm{TPR}}
\newcommand{\FPR}{\mathrm{FPR}}
\newcommand{\FN}{\mathrm{FN}}

\DeclareMathOperator*{\ENTROPY}{Entropy}

\begin{document}
\title{Towards Truly Unsupervised Evaluation of Feature Selection -- Extended Version}
\titlerunning{Towards Truly Unsupervised Evaluation of Feature Selection}
%
%
\author{Hafiz Saud Arshad\inst{1}\orcidID{0009-0003-2107-5589} \and
Muhammad Rajabinasab\inst{1}\orcidID{0009-0006-7045-3998} \and
Arthur Zimek\inst{1}\orcidID{0000-0001-7713-4208}}
\authorrunning{H.~S.~Arshad et al.}
%
\institute{University of Southern Denmark, Odense, Denmark
\email{\{saudarshad,rajabinasab,zimek\}@imada.sdu.dk}
}
\maketitle              
\begin{abstract}
Feature selection is one of the most important and fundamental tasks in data mining, tackled by a family of methods with an established set of evaluation techniques to measure the quality of a specific method. Most of the methods commonly used for the unsupervised evaluation of feature selection algorithms suffer from critical design flaws which question their unsupervised nature. In this paper, we provide a critical discussion on the established allegedly unsupervised evaluation techniques, and shed light on the reasons why they are not truly unsupervised but, at best, supervised evaluation under an unsupervised downstream task. We also propose a novel, truly unsupervised evaluation framework to measure the quality of the feature selection algorithms without any form of information about the labels. The proposed framework utilizes unsupervised Principal Component Analysis, and optimal transport to measure the quality of the feature selection methods in a truly unsupervised manner.

\end{abstract}
\section{Introduction}
The curse of dimensionality is one of the fundamental challenges in machine learning and data mining. As dimensionality increases, both computational complexity and the difficulty of downstream tasks grow considerably. High dimensionality negatively impacts a wide range of applications, including but not limited to Nearest Neighbor search~\cite{DBLP:conf/ssdbm/HouleKKSZ10,RAJABINASAB2025126254}, classification~\cite{DBLP:journals/tifs/AmsalegBBEFHRN21}, clustering~\cite{DBLP:conf/sisap/OkkelsTAZS25}, and outlier detection~\cite{DBLP:conf/sdm/Anderberg0CHMRZ24,rajabinasab2026randomizedpcaforestunsupervised,rajabinasab2025hdsvdd}, among others. Dimensionality reduction consists of many different families of techniques, ranging from subspace learning~\cite{rajabinasab2025hdsvdd} to feature extraction~\cite{asadi2021face}. Among dimensionality reduction methods, Feature Selection remains particularly important because it preserves the semantics of the original dimensions and therefore maintains interpretability and explainability.

Feature selection is a fundamental machine learning and data mining task. Given a set of initial features $F = \{f_1, f_2, \dots, f_n\}$, the problem of feature selection is defined as finding the optimal subset $S^* \subseteq F$ that maximizes a criterion function $J(S)$ (e.g., the predictive power or the information content of the subset). Formally, this can be expressed as the optimization problem ${S^* = \arg \max_{S \subseteq X} J(S)}$. It can be set to be subject to the constraint ${|S| = k < n}$ or with a penalty term that encourages sparsity. The objective is to identify the most relevant features while eliminating redundant or irrelevant dimensions, thereby improving model generalization and reducing computational costs~\cite{guyon2003introduction,li2017feature}. Feature selection can be conducted in both supervised and unsupervised settings. In supervised feature selection, information about class labels is used to facilitate the identification of informative features. In the unsupervised setting, the feature selection process is based on the intrinsic information of the features themselves (e.g., correlation and variance)~\cite{cai2018feature}.

Similar to dimensionality reduction, feature selection addresses the \emph{curse of dimensionality}, the exponential expansion of the data space as the number of features increases, causing data to become so sparse that statistical distance and density metrics lose their meaningful distinction, leading to a degradation in the performance of many downstream machine learning and data mining tasks~\cite{DBLP:journals/tkde/FrancoisWV07,DBLP:journals/sadm/ZimekSK12}. Unlike dimensionality reduction methods, feature selection offers an important advantage: keeping the explainability properties of the original feature space while reducing the dimensionality. This is why, despite the existence of many successful methods for dimensionality reduction such as Principal Component Analysis (PCA), Autoencoders~\cite{bourlard1988}, t-distributed Stochastic Neighbor Embedding (t-SNE)~\cite{vandermaaten2008}, and Uniform Manifold Approximation and Projection (UMAP)~\cite{mcinnes2018}, feature selection methods are still important for tackling the \emph{curse of dimensionality}~\cite{guyon2003introduction}.

There are established evaluation techniques for measuring the quality of feature selection algorithms. In most cases, a combination of supervised and unsupervised evaluation is used which measures the quality of a feature selection method with respect to the performance in conducting a downstream task. There is however an important problem concerned with the unsupervised evaluation part, and that is the fact that the process is done by conducting clustering (usually using $k$-means) on the data representation using selected features, and checking the clustering accuracy with respect to the ground-truth labels. This is clearly problematic, as using any information about the labels is in contrast with the alleged unsupervised nature which we expect from the evaluation, rendering the current established so-called unsupervised way of evaluation, at best, as \emph{supervised evaluation under an unsupervised downstream task}. Even not using the labels, and just using the number of distinct classes to select the number of clusters, should still be considered a violation of the unsupervised setting. Unsupervised feature selection methods require a truly unsupervised evaluation technique, and in many practical scenarios, ground-truth labels might actually not be available.

There are other evaluation metrics and methods proposed for the evaluation of feature selection algorithms. These techniques offer ways of evaluating the entire feature selection process, instead of a single point~\cite{rajabinasab2024}, evaluating the amount of dimensionality reduction and the quality of feature selection at once~\cite{mostert2021feature}, or the evaluation of the stability of feature selection algorithms~\cite{nogueira2017stability,rajabinasab2024}, or of model-agnostic evaluation~\cite{rajabinasab2025interdataset1}. Only the last one can be considered an unsupervised metric, as it does not use the labels or any external information about the labels in the calculation of the metric.

In this paper, we propose a novel method for the \emph{truly} unsupervised evaluation of feature selection algorithms, utilizing dimensionality reduction and inter-dataset similarity to measure the quality of the feature selection process. In this evaluation method, we take advantage of PCA as an informative unsupervised dimensionality reduction technique. We evaluate the performance of a feature selection algorithm $A$ in selecting $f$ features by calculating the inter-dataset similarity metric $\mathit{IDS}$ between the subsets constituted by selecting $f$ features using algorithm $A$, and the first $f$ principal components. Inter-dataset similarity metrics measure the distributional similarities or differences between two datasets. There are various inter-dataset similarity measures~\cite{rajabinasab2025interdataset1}, especially in the field of optimal transport (OT)~\cite{peyre2019computational}. Metrics such as Earth Mover's Distance (EMD)~\cite{rubner2000earth}, Wasserstein Distance~\cite{villani2009optimal}, Sliced Wasserstein Distance~\cite{nguyen2025lightspeed}, and Sinkhorn~\cite{cuturi2013sinkhorn} can give us useful insights on how similar two datasets are. Measuring the inter-dataset similarity between the two aforementioned subsets of the data allows us to rank the feature selection algorithms based on how closely to PCA they perform. This process hence favors a feature selection algorithm yielding close performance figures to PCA without compromising the explainability as a notion for unsupervised evaluation.

The remainder of this paper is structured as follows: In Section~\ref{sec:fseval}, we review the existing measures for evaluating feature selection algorithms. In Section~\ref{sec:prop}, we showcase a detailed description of the proposed unsupervised evaluation framework. In Section~\ref{sec:exp}, we present experimental results. In Section~\ref{sec:conc}, we conclude the paper by presenting a summary, highlighting limitations, and outlining possible directions for future research. 

\section{Related Work} \label{sec:fseval}
Feature selection algorithms are primarily evaluated based on their ability to enhance (or maintain) the performance of downstream machine learning tasks while simultaneously reducing the dimensionality. Evaluation methods for feature selection can be divided into two main categories: (i) performance assessment on downstream supervised or unsupervised tasks, typically using metrics computed for a fixed number of selected features, and (ii)  measures that capture broader properties such as overall performance trends, model-agnostic quality, fitness gain, or stability.

\subsection{Evaluation Based on Downstream Tasks}

The most common evaluation approach involves measuring how feature selection affects the quality of a subsequent machine learning task, most frequently classification (supervised) or clustering (unsupervised). These evaluations are usually point-based, i.e., they report performance achieved with a specific number of selected features.

\subsubsection{Supervised Evaluation:}

In supervised settings, the effectiveness of a feature selection method is typically quantified by training one or more classifiers on the reduced feature set and calculating standard classification performance measures.

\paragraph{Accuracy (ACC):}
Accuracy is defined as the proportion of correctly classified instances:
\begin{equation}
    \ACC = \frac{\TP + \TN}{\TP + \TN + \FP + \FN}
\end{equation}
where $\TP$, $\TN$, $\FP$, and $\FN$ denote the number of true positives, true negatives, false positives, and false negatives, respectively.

\paragraph{Area Under the ROC Curve (AUC):}
The area under the receiver operating characteristic curve quantifies a classifier’s ability to discriminate between classes across all decision thresholds:
\begin{equation}
    \AUC = \int_{0}^{1} \TPR (\FPR) \, d\FPR
\end{equation}
where $\TPR$ is the true positive rate and $\FPR$ is the false positive rate. Values of $\AUC$ close to 1 indicate excellent discriminative power, while $\AUC = 0.5$ corresponds to random guessing.

Supervised metrics are used widely for the evaluation of feature selection methods. However, their values strongly depend on the choice of classifier, hyperparameter settings, and the dataset characteristics. They can be deceptive, as classifiers might still perform reasonably well even on a poorly (or randomly) selected subset of features~\cite{rajabinasab2026worserandomimportancebaseline}.

\subsubsection{Unsupervised Evaluation:}

In the so-called unsupervised evaluation of feature selection, performance is frequently evaluated by applying a clustering algorithm (usually k-means) on the selected features and comparing the resulting clustering labels against ground-truth class labels using external validation measures, i.e., these evaluation measures actually are not unsupervised as such, but only use an unsupervised downstream task while employing supervised evaluation.

\paragraph{Clustering Accuracy (CLSACC)}
Clustering accuracy measures the agreement between predicted cluster labels and true class labels after finding an optimal label correspondence:
\begin{equation}
    \CLACC = \frac{1}{n} \sum_{i=1}^{n} \delta\bigl(c_i, \text{map}(g_i)\bigr)
\end{equation}
where $n$ is the number of instances, $c_i$ and $g_i$ are the predicted cluster and true class labels for instance $i$, respectively, $\text{map}(\cdot)$ is the optimal assignment obtained via the Hungarian method~\cite{kuhn1955hungarian}, and $\delta(\cdot,\cdot)$ is the Kronecker delta function~\cite{Bishop2006}.

\paragraph{Normalized Mutual Information (NMI):}
Normalized mutual information quantifies the shared information between the predicted clustering $C$ and the true partitioning indicated by the ground-truth labels $L$, normalized by their entropies:
\begin{equation}
    \NMI = \frac{\MI(L, C)}{\max\bigl(\ENTROPY(L), \ENTROPY(C)\bigr)}
\end{equation}
where $\MI(\cdot,\cdot)$ denotes mutual information. The measure ranges from 0 (independent) to 1 (perfect agreement).

These external clustering metrics, while informative, still rely on ground-truth labels, which is in contrast with the unsupervised nature of the downstream task. If we aim to apply an unsupervised feature selection algorithm on a dataset which does not have any labels, this would be rendered ineffective as much as the supervised evaluation.

\subsection{Specialized Evaluation Measures}

Several specialized metrics have been proposed to overcome limitations of single-point and downstream task-dependent evaluations by providing more holistic, model-agnostic, or robustness-oriented insights.

\subsubsection{FSDEM Score:}

Feature Selection Dynamic Evaluation Metric (FSDEM)~\cite{rajabinasab2024} aggregates performance across varying numbers of selected features. Let $d$ be the total number of features, $P(f)$ be the value of a downstream performance measure obtained using exactly $f$ selected features, and $g(x)$ be a continuous function approximated based on the observed pairs $\{(f, P(f))\}_{f=1}^{d}$. The FSDEM score is then defined as the normalized area under this curve over a chosen interval $[a,b]$:
\begin{equation}
    \FSDEM = \frac{1}{b-a} \int_{a}^{b} g(x) \, dx.
\end{equation}
When the underlying performance measure $P$ is bounded in $[0,1]$, so is $\FSDEM$, with values closer to 1 indicating consistently good performance. FSDEM allows approximating the overall performance of the entire feature selection process based on a number of observations, leading to a more holistic overview compared to the common single-point evaluation.

\subsubsection{Average Angle Difference (AAD):}

The Average Angle Difference~\cite{rajabinasab2025interdataset1}, also formally denoted as $\overline{\Delta\theta}$, provides a model-agnostic assessment of feature selection quality by quantifying the average change in the direction of the eigenvector corresponding to the first principal component when non-selected features are removed. Let $F$ be the set of selected features and $F^C$ its complement. For each feature $f \in F^C$, construct the modified dataset $\mathbf{D}_{f^0}$ by setting all values of feature $f$ to zero. The AAD is then calculated as:
\begin{equation}
    AAD = \frac{1}{|F^C|} \sum_{f \in F^C} \Delta\theta\bigl(\mathbf{D}, \mathbf{D}_{f^0}\bigr),
\end{equation}
where $\Delta\theta$ denotes the angle difference between the eigenvectors corresponding to the first principal components. Smaller values of $\overline{\Delta\theta}$ (closer to 0) indicate that the removed features were largely redundant or uninformative, hence affecting the first and most important principal component the least.

\section{Unsupervised Evaluation of Feature Selection} \label{sec:prop}
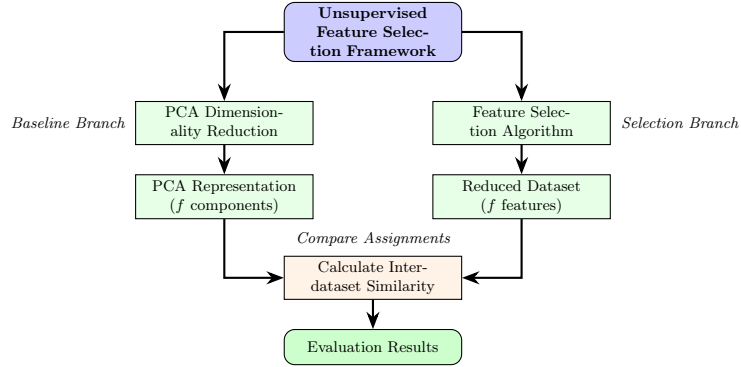
\begin{figure}[tb]
    \centering
    \begin{tikzpicture}[
        node distance=0.6cm and 0.5cm, 
        scale=0.65,
        transform shape,
        block/.style={rectangle, draw, fill=blue!10, text width=3.4cm, text centered, rounded corners, minimum height=2.2em, inner sep=3pt},
        proc/.style={rectangle, draw, fill=green!10, text width=3.4cm, text centered, minimum height=2.2em, inner sep=3pt},
        arrow/.style={-Stealth, thick},
        label_node/.style={font=\small\itshape}
    ]
        \node (start) [block, fill=blue!20] {\textbf{Unsupervised Feature Selection Framework}};
        
        \node (dr) [proc, below left=0.8cm and -0.6cm of start] {PCA Dimensionality Reduction};
        \node (clustering_dr) [proc, below=of dr] {PCA Representation ($f$ components)};
        
        \node (fs) [proc, below right=0.8cm and -0.6cm of start] {Feature Selection Algorithm};
        \node (clustering_fs) [proc, below=of fs] {Reduced Dataset ($f$ features)};
        
        \node (metrics) [proc, fill=orange!10] at ($(clustering_dr)!0.5!(clustering_fs) + (0,-1.65)$) {Calculate Inter-dataset Similarity};
        \node (output) [block, below=0.6cm of metrics, fill=green!20] {Evaluation Results};
        
        \draw [arrow] (start) -| (dr);
        \draw [arrow] (start) -| (fs);
        
        \draw [arrow] (dr) -- (clustering_dr);
        \draw [arrow] (fs) -- (clustering_fs);
        
        \draw [arrow] (clustering_dr) |- (metrics);
        \draw [arrow] (clustering_fs) |- (metrics);
        \draw [arrow] (metrics) -- (output);
        
        \node [label_node, left=0.1cm of dr] {Baseline Branch};
        \node [label_node, right=0.1cm of fs] {Selection Branch};
        
        \node [label_node, above=0.08cm of metrics] {Compare Assignments};
    \end{tikzpicture}
    \caption{Proposed method for truly unsupervised evaluation of feature selection.}
    \label{fg:framework}
\end{figure}

The core idea behind our framework is to decouple the evaluation from the original ground-truth labels. The overall structure of the method is presented in Fig.~\ref{fg:framework}. The method operates as follows: We apply PCA for effective dimensionality reduction and select top $f$ principal components. 
To evaluate the result of some feature selection algorithm, selecting $f$ features, we calculate an inter-dataset similarity metric $\mathit{IDS}$ for the two representations of the data resulting from PCA and from some feature selection method. The similarity between the reduced datasets with $f$ selected features and the PCA-based representation with $f$ components yields the final evaluation.

The objective is to identify feature selection algorithms that select features in a way that they are as close as possible to the representation obtained from PCA. The better a subset preserves the structure of the PCA representation, the higher it ranks.

The proposed framework constitutes a truly unsupervised approach for evaluating the quality of feature selection algorithms. The use of PCA as a baseline allows us to favor feature selection methods which select features which best resemble the principal components, aiming for a similar downstream performance figure while maintaining the explainability. The unsupervised nature of the proposed framework addresses the existing gap in the evaluation of feature selection algorithms by facilitating the truly unsupervised evaluation of the feature selection methods.

While we focus on the implementation of the framework using PCA, other dimensionality reduction methods could be used as well, depending on the desired bias of the quality measure. We experiment with a selection of different dataset similarity measures. Earth Mover's Distance ({OT\_EMD2}) \cite{DBLP:journals/tog/BonneelPPH11} measures the minimum cost of transforming one distribution into another by moving probability mass. Entropic-Regula\-rized Sinkhorn Distance ({OT\_SINKHORN2}) \cite{DBLP:conf/nips/Cuturi13} is a GPU-friendly approximation of EMD. Gromov-Wasserstein Distance ({OT\_GW2}) \cite{GromoveWasserstein2} compares distributions living in incomparable metric spaces by matching their internal pairwise geometry rather than point-to-point correspondences. And finally, Sliced Wasserstein Distance ({OT\_Sliced\_SW}) \cite{DBLP:journals/jmiv/BonneelRPP15} approximates the Wasserstein distance by averaging one-dimensional projections of the distributions, yielding a scalable but lower-resolution similarity estimate. 

As all the values reflected by optimal transport are a measure of distance or dissimilarity, we transform them into a similarity measure:

\begin{equation}
\OT_\text{sim} = \frac{1}{\OT} \quad \text{if } \OT \neq 0 \text{ else } \OT = +\infty 
\end{equation}

where $\OT$ is the optimal transport cost. As the distances are theoretically not bounded, we use $+\infty$ as an indicator of maximum possible similarity which only happens when the distributions are exactly the same (e.g., passing the same dataset twice).

\section{Experiments and Analyses} \label{sec:exp}
We perform our experiments using the \texttt{FSEVAL} benchmarking suite~\cite{rajabinasab2026fsevalfeatureselectionevaluation}, a specialized framework designed for the extensive and comprehensive evaluation of feature selection algorithms. Below, we list the experimental setup, results and analysis.

\subsection{Experimental Setup} \label{subsec:es}

The benchmark consists of 8 publicly available high-dimensional datasets commonly used in the unsupervised feature selection literature~\cite{li2017feature}. This collection spans biomedical data and other high-dimensional benchmarks, representing a variety of applications. These datasets include COIL20, Isolet, ORL, lung, lung\_discrete, warpAR10P, warpPIE10P, and Yale.

\subsubsection{Unsupervised Feature Selection Methods} \label{subsec:ufs}
Among the classical approaches, variance-based~\cite{he2005laplacian} and correlation-based methods~\cite{hall1999correlation} operate directly on feature statistics without any learning component. Laplacian Score (LS)~\cite{laplacianscore} is a forward-filter method that operates in an unsupervised manner as part of the broader spectral feature selection framework~\cite{spectral}, with computational complexity of $C_{\mathrm{LS}} = dn^{2}$, where $d$ denotes the number of features and $n$ the number of samples. 

On the more recent end, Subspace Clustering-based Feature Selection (SCFS) \cite{DBLP:journals/eaai/ParsaZG20} integrates subspace learning, clustering, 
and sparse learning via a self-expressive model combined with regularized regression to 
sparsely capture feature-cluster correlations. The Variance-Covariance Subspace Distance approach (VCSDFS) leverages inter-feature correlations by identifying feature subsets whose variance-covariance matrices exhibit minimal norm properties~\cite{DBLP:journals/nn/KaramiSTMV23}. 

We evaluate five feature selection methods and augment them with the Random baseline~\cite{rajabinasab2026worserandomimportancebaseline} that assigns uniform random importance scores to study how the proposed metrics are able to discriminate the random behavior from other feature selection approaches. Hyperparameters follow the default or suggested values, e.g., Laplacian Score uses $k=5$ for the nearest neighbor graph, VCSDFS uses $\rho=0.5$ with $30$ iterations, SCFS uses $\alpha=\beta=1.0$.

\subsubsection{Evaluation Measures}
On the evaluation end, we compare nine feature selection evaluation metrics, organized into four groups:
\begin{itemize}
    \item \textbf{Supervised:} These metrics use explicit supervisory knowledge e.g., ground-truth class labels. We use Classification Accuracy (ACC) and Area under the ROC Curve (AUC), computed with a Random Forest classifier (100 trees) under the stratified 5-fold cross validation, average over 5 repetitions (different splits).
    \item \textbf{Pseudo-Unsupervised:} These metrics are based on clustering as an unsupervised downstream task, but are still measured with respect to the ground-truth labels, including Clustering Accuracy (CLSACC) and Normalized Mutual Information (NMI), obtained from $k$-means with the number of clusters set to the number of ground-truth classes, averaged over 10 different random initializations. 
    \item \textbf{Model Agnostic:} The Average Angle Difference (AAD) \cite{rajabinasab2024} does not use any label information and operates on the average change in the angle of the eigenvector corresponding to the first principle component caused by the removal of unselected features.
    \item \textbf{Proposed (Truly Unsupervised):} We create four instantiations of our framework, differing in the choice of the inter-dataset similarity metric while comparing the optimal transport between the subspace of selected features against the PCA-reduced representation. As discussed in Section~\ref{sec:prop}, we are using the Earth Mover's Distance (OT\_EMD2) \cite{DBLP:journals/tog/BonneelPPH11}, Entro\-pic-Regula\-rized Sinkhorn Distance (OT\_SINKHORN2) \cite{DBLP:conf/nips/Cuturi13}, Gromov-Wasser\-stein Distance (OT\_GW2) \cite{GromoveWasserstein2}, and Sliced Wasserstein Distance (OT\_SLICED\_SW) \cite{DBLP:journals/jmiv/BonneelRPP15}.
    
\end{itemize}

Following the dynamic-evaluation philosophy of FSDEM \cite{rajabinasab2024}, we assess each method at 10 feature fractions in low-percentage regime, \textit{0-5 Percent} with p $\in$ \{0.005, 0.010, 0.015, \dots , 0.05\}, capturing the low-feature setting most relevant to high-dimensional data.

\subsubsection{Rankings, Repetitions, and Selection}
For each 
\textit{(dataset, feature selection method, percentage)} configuration, we proceed as follows. The feature selection method assigns a score to every feature; scores are sorted in descending order, and the top-$f$ features are selected where $f$ is determined by the percentage $p$ following the FSDEM schedule~\cite{rajabinasab2024}. Stochastic methods (here, only the $Random$ baseline) are run $10$ times and their resulting metric values are averaged; while deterministic methods are run once. We evaluate the selected set of features by each feature selection method using all 9 metrics.

\begin{figure}[tb] 
    \centering
    \includegraphics[width=0.9\linewidth]{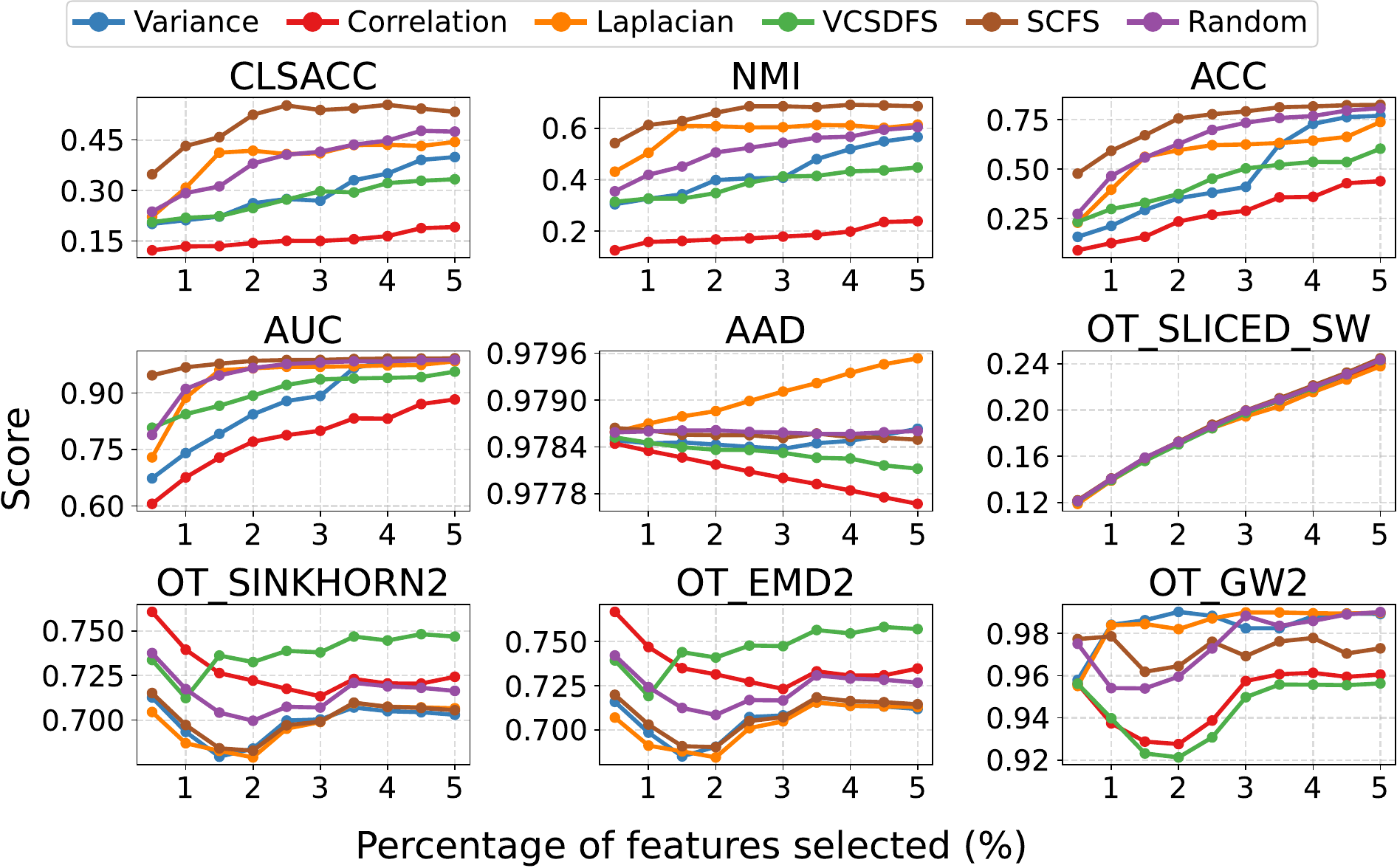}
    \caption{Performance curves of the nine evaluation metrics on \textit{Isolet} as a function of the percentage of features selected. For all measures, higher is better (distances are converted to similarities for OT-measures).}
    \label{fig:Isolet_5Percent}
\end{figure}

\subsection{Results}
\begin{figure}[tb] 
    \centering
    \includegraphics[width=0.7\linewidth]{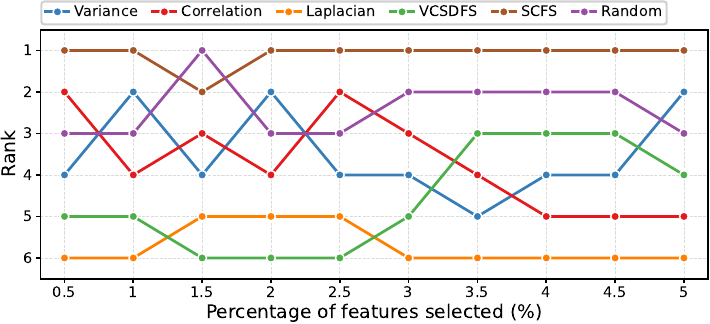}
    \caption{Ranking produced by \textit{Ot\_Sliced\_SW} for \textit{Isolet} dataset as a function of the percentage of features selected. $1$ represents highest rank and $6$ represents lowest rank.}
    \label{fig:Isolet_Sliced}
\end{figure}

Figure~\ref{fig:Isolet_5Percent} shows that \textit{OT\_GW2} produces rankings partly aligning with the supervised block by placing \textit{VCSDFS} and \textit{Correlation} ranked lowest and \textit{Laplacian} ranked highest. Although it places \textit{SCFS} fourth and \textit{Variance} second, misplaced, it still shows an overall agreement. Such disagreements are observed within Supervised block as well across different datasets. While \textit{OT\_SINKHORN2} and \textit{OT\_EMD2} produce the ranking largely inverted with respect to the supervised block, rating the two weakest selectors best. Here, we highlight that the local disruption between $1.5\%$ and $3.5\%$ is caused by the per-budget cost matrix normalization performed on the OT-metrics as a standard step: the unnormalized distance grows monotonically, while its denominator grows faster. Hence, we recommend readers not to compare OT scores across different percentages of selected features, rather observe the ranking order. In case of \textit{OT\_SLICED\_SW}, the resolution degrades. Figure~\ref{fig:Isolet_Sliced} shows the ranking produced by \textit{OT\_SLICED\_SW} for the six feature selection methods and we see traces of agreement e.g., \textit{Random} and \textit{SCFS} are ranked top. Performance curves for the remaining datasets are provided in Appendix~\ref{app:curves}, and the corresponding \textit{OT\_SLICED\_SW} rankings in Appendix~\ref{app:sliced}.

These observations indicate that the proposed framework carries predictive validity: the proposed (label-free) metrics order selectors in a way that anticipates their label-based downstream performance. This should not be read as a claim of constructing \textit{validity} nor as measuring our pipeline's efficacy by its agreement with supervised metrics. Agreement should rather be seen as the downstream-task performance (here, classification) of the selected features that a practitioner would obtain by choosing methods according to the ranking provided by the proposed measure. Furthermore, a disagreement should not be seen as declaring a metric ineffective. We should not expect them to produce a single shared ordering of the selectors; rather, examining where the rankings agree and where they diverge is itself informative about the evaluation landscape. The metrics provide us with information about a specific aspect of the quality of the feature selection process. Our framework measures similarity to the first $f$ principal components which is in particular important as there is no loss of explainability imposed by feature selection, compared to the case of PCA, where the actual features are replaced by principal components.

\begin{figure}[tb]
    \centering
    \includegraphics[width=0.92\linewidth]{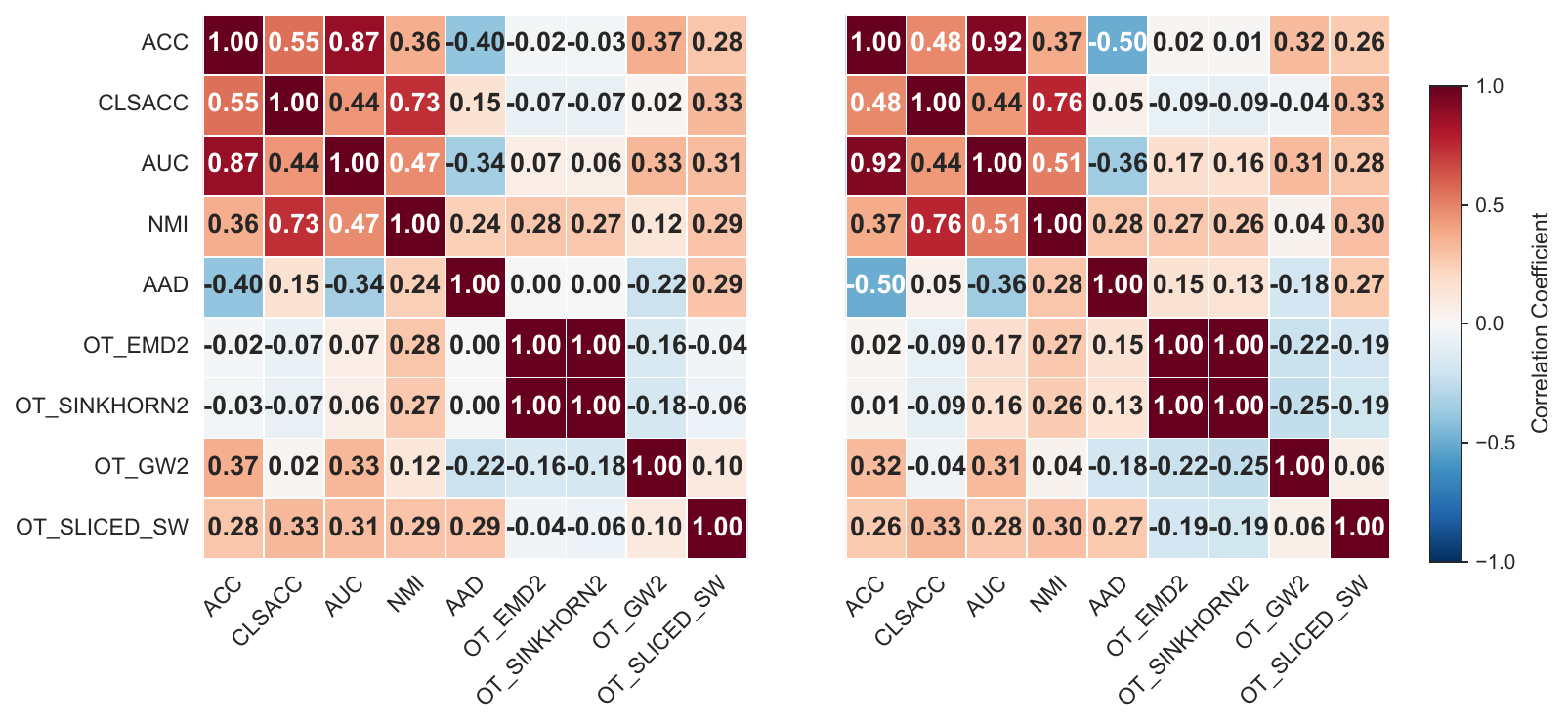}
    \caption {Pairwise correlation between 9 evaluation metrics for the percentage of features selected in the range (0.5\%--5\%). The left plot reports Pearson correlations; and the right plot reports Spearman. Red cells mark metric-pair that ranks the feature selection methods consistently in the same direction, blue cells represent a pair that ranks them in opposite directions.}
    \label{fig:metric_correlations}
\end{figure}

Figure~\ref{fig:metric_correlations} reports the pairwise correlations between evaluation metrics, separately under Pearson and Spearman correlation. It is evident that \textit{OT\_GW2} and \textit{OT\_SLICED\_SW} exhibit positive correlation with respect to traditional metrics, especially the supervised block(Pearson coefficients: $0.29-0.37$; Spearman coefficients: $0.26-0.33$), While \textit{OT\_EMD2} and \textit{OT\_Sinkhorn2} exhibit opposite patterns. Among model-agnostic alternatives, \textit{AAD} correlates positively but relatively weakly with the supervised block than the proposed \textit{optimal transport}-based metrics, indicating that the proposed framework tracks label-based downstream performance competitively without explicit supervision.

\section{Conclusions, Limitations, and Future Work} \label{sec:conc}

This study takes a first step toward a broader question: whether feature selection can be evaluated without relying on information that the feature selection process itself is not assumed to have access to. We have approached this question by replacing label-dependent evaluation with an unsupervised comparison between the selected feature space and a reference representation derived from the data themselves. The results provide encouraging evidence for the feasibility of this perspective, while also revealing several methodological and empirical challenges that remain to be addressed. The following subsections summarize the main conclusions, discuss the limitations of the current framework, and outline directions for developing a more comprehensive and robust approach to truly unsupervised evaluation.

\paragraph{\textbf{Conclusions.}} In this paper, we argued that the established procedures for unsupervised evaluation of feature selection are not truly unsupervised. The procedures commonly used in the literature as unsupervised evaluation in fact rely on ground-truth labels. Hence, these techniques are technically supervised, merely using an unsupervised downstream task. To the best of our knowledge, only one truly unsupervised and model-agnostic metric currently exists for the evaluation of feature selection methods~\cite{rajabinasab2025interdataset1}.

We proposed a truly unsupervised evaluation framework that, instead of exploiting ground-truth labels, evaluates a feature selection algorithm by measuring the \emph{optimal transport} distance between its selected feature subset and a PCA-based reference subspace of the same size. Across eight high-dimensional datasets, the proposed metrics showed  consistent correlations with established evaluation metrics under both correlation measures.

Our results also demonstrate that the choice of the underlying inter-dataset similarity measure has a substantial effect on the insights obtained from the evaluation. Importantly, a lower correlation with classification-based performance should not necessarily be interpreted as evidence that a metric provides a false or uninformative signal. Rather, it may indicate that the metric captures aspects of feature selection quality that are different from those reflected in downstream classification performance. This observation reinforces the importance of studying unsupervised evaluation metrics not merely according to their agreement with supervised performance, but also according to the properties of the selected representations that they actually measure.

\paragraph{\textbf{Limitations.}} Despite providing preliminary steps toward truly unsupervised evaluation of feature selection, the proposed framework has several important limitations.

First, optimal transport methods can be computationally expensive, particularly for large datasets. Computationally efficient approximations, such as Sinkhorn-based methods~\cite{cuturi2013sinkhorn}, provide a practical alternative but may introduce stochasticity. This is particularly problematic for an evaluation metric, for which deterministic behavior is desirable. Given the same feature selection process and dataset, repeated evaluations should ideally produce identical results. Such stability is important when metric values are subsequently used for comparisons, statistical analysis, or ranking of feature selection methods.

Second, the use of PCA as the reference mechanism imposes a structural limitation on the framework. The number of principal components that can be obtained from PCA is bounded by $\min(n,d)$, where $n$ is the number of instances and $d$ is the number of features. Consequently, the proposed evaluation is either restricted to datasets satisfying $n \geq d$, or limited to a partial view of the feature selection process when the number of selected features exceeds this bound. Ideally, an unsupervised evaluation metric should be applicable to datasets regardless of the relationship between their numbers of instances and features, and should be capable of evaluating the selection process across its full range of dimensionality reductions.

Finally, the empirical scope of this study is limited. The experiments cover a relatively small number of datasets and feature selection methods, and the observed relationships may therefore not generalize to the broader space of feature selection problems. Moreover, the current analysis primarily examines aggregate correlations rather than investigating how these relationships vary across individual datasets or different levels of dimensionality reduction.

\paragraph{\textbf{Future Work and Broader Impact.}} The limitations above suggest several directions for future research. First, the proposed framework should be studied at a finer granularity. In particular, correlations should be examined separately for individual datasets and for different levels of dimensionality reduction. Repeated experiments should also be conducted to quantify and, where possible, eliminate the effects of stochasticity. A systematic comparison of different optimal transport formulations and their underlying design choices is likewise necessary to understand why they produce different evaluation signals. Such analyses would provide a more complete picture of the behavior of the proposed framework and help establish reliable recommendations for its use.

Second, the reliance on PCA as the unsupervised reference could be relaxed. Other dimensionality reduction techniques, such as autoencoders, could provide alternative reference representations and potentially overcome the dimensionality constraints of PCA while also allowing nonlinear structures to be captured. This could further enable a more detailed analysis of method rankings using Critical Difference diagrams based on Standard rank statistics~\cite{demsar2006statistical} and MARS~\cite{rajabinasab2026marsmagnitudeawarerankstatistics}. More generally, alternative dimensionality reduction techniques could transform the current approach into a dynamically guided evaluation framework. For example, multiple dimensionality reduction candidates could be generated and evaluated using internal dimensionality-reduction quality measures~\cite{van2009dimensionality}, with the best-performing representation for a given dataset subsequently used as the reference for feature selection evaluation.

Third, the raw values produced by the proposed metrics warrant further investigation. It should be determined whether their distributions have a meaningful interpretation and whether a correction-for-chance adjustment is necessary~\cite{hubert1985comparing}. Such an adjustment could improve the interpretability and comparability of metric values across datasets and experimental settings.

Fourth, the properties of individual optimal transport measures provide an opportunity to simplify the framework. In particular, the Gromov-Wasserstein distance~\cite{GromoveWasserstein2}, used in \texttt{OT\_GW2}, can compare distributions defined on spaces of different dimensionality. This property could potentially eliminate the need for a reference subspace altogether, allowing the selected feature subset to be compared directly with the full-dimensional dataset.

Finally, the recurring loss of resolution observed for \texttt{OT\_SLICED\_SW} deserves a systematic investigation. Understanding why this behavior occurs, under which dataset characteristics it emerges, and which alternative inter-dataset similarity measures reproduce or avoid it could provide important insights into the design of future unsupervised evaluation metrics.

Taken together, these directions position the present study as a starting point rather than a final characterization of unsupervised feature selection evaluation. The limited number of datasets and methods, together with the methodological constraints imposed by optimal transport and PCA, mean that the current results should primarily be interpreted as a proof of concept. We therefore do not intend the observed correlations to constitute a definitive verdict on the superiority of the proposed metrics or on the quality of existing feature selection methods. Instead, the broader aim is to demonstrate the feasibility of evaluating feature selection without relying on ground-truth labels and to motivate a more rigorous, granular, and reproducible investigation of truly unsupervised evaluation in future work.

\begin{credits}
\subsubsection{\ackname}This study was funded jointly by The Independent Research Fund Denmark (DFF) in the project ``CLEVR: Clustering Evaluation Revisited' and ``PREPARE: Personalized Risk Estimation and Prevention of Cardiovascular Disease''.

\end{credits}

\bibliographystyle{splncs04}
\bibliography{condensed_dblp}

\clearpage
\appendix
\makeatletter
\setlength{\@fptop}{0pt}
\setlength{\@fpsep}{10pt}
\setlength{\@fpbot}{0pt plus 1fil}
\makeatother
\section*{Appendix}
\setcounter{section}{1}
\subsection{Per-Dataset Performance Curves} \label{app:curves}
\begin{figure}[!htbp]
    \centering
    \includegraphics[width=0.9\linewidth]{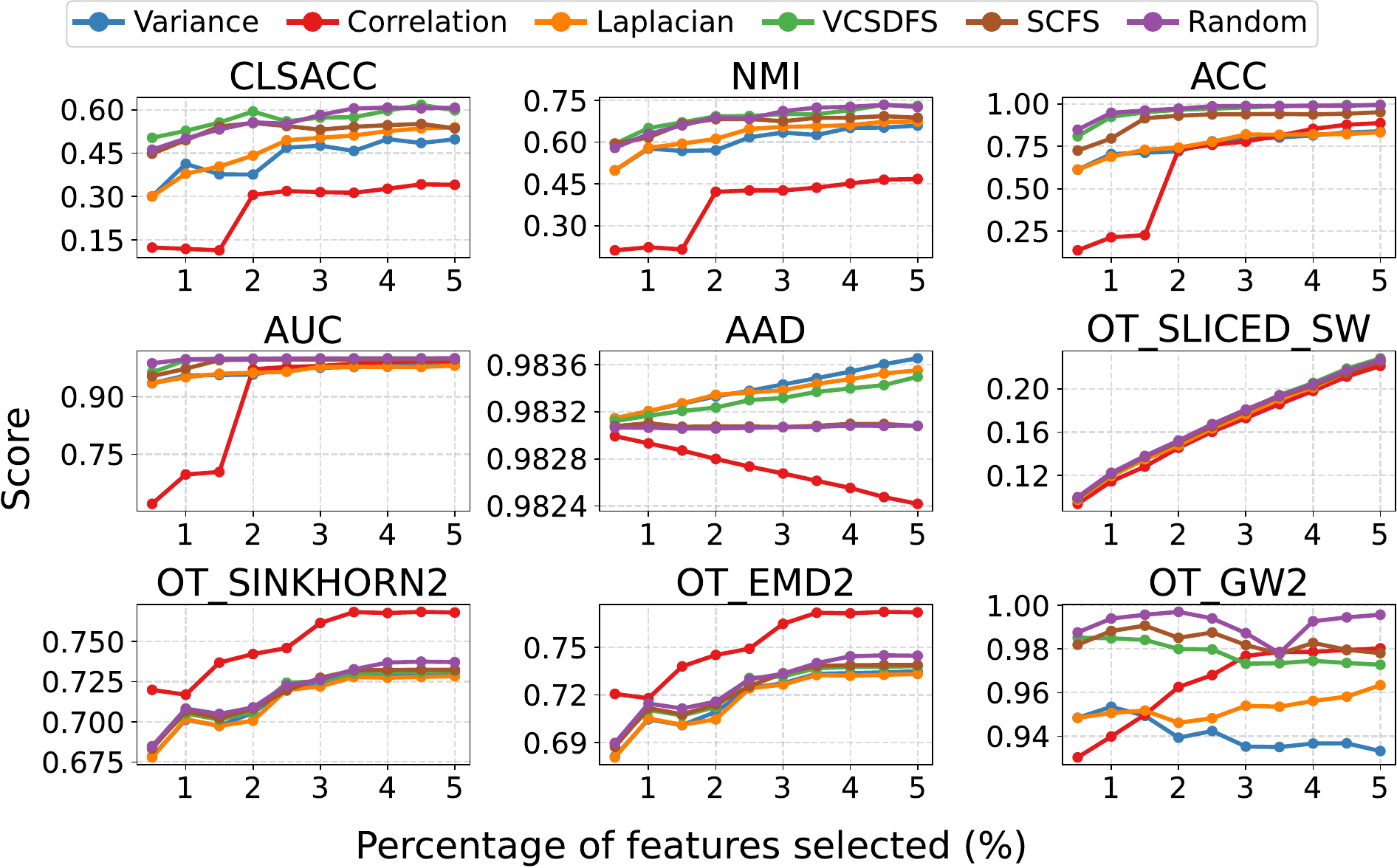}
    \caption{Performance curves of the nine evaluation metrics on \textit{COIL20} as a function of the percentage of features selected.}
    \label{fig:COIL20_5Percent}
\end{figure}

\begin{figure}[!htbp]
    \centering
    \includegraphics[width=0.9\linewidth]{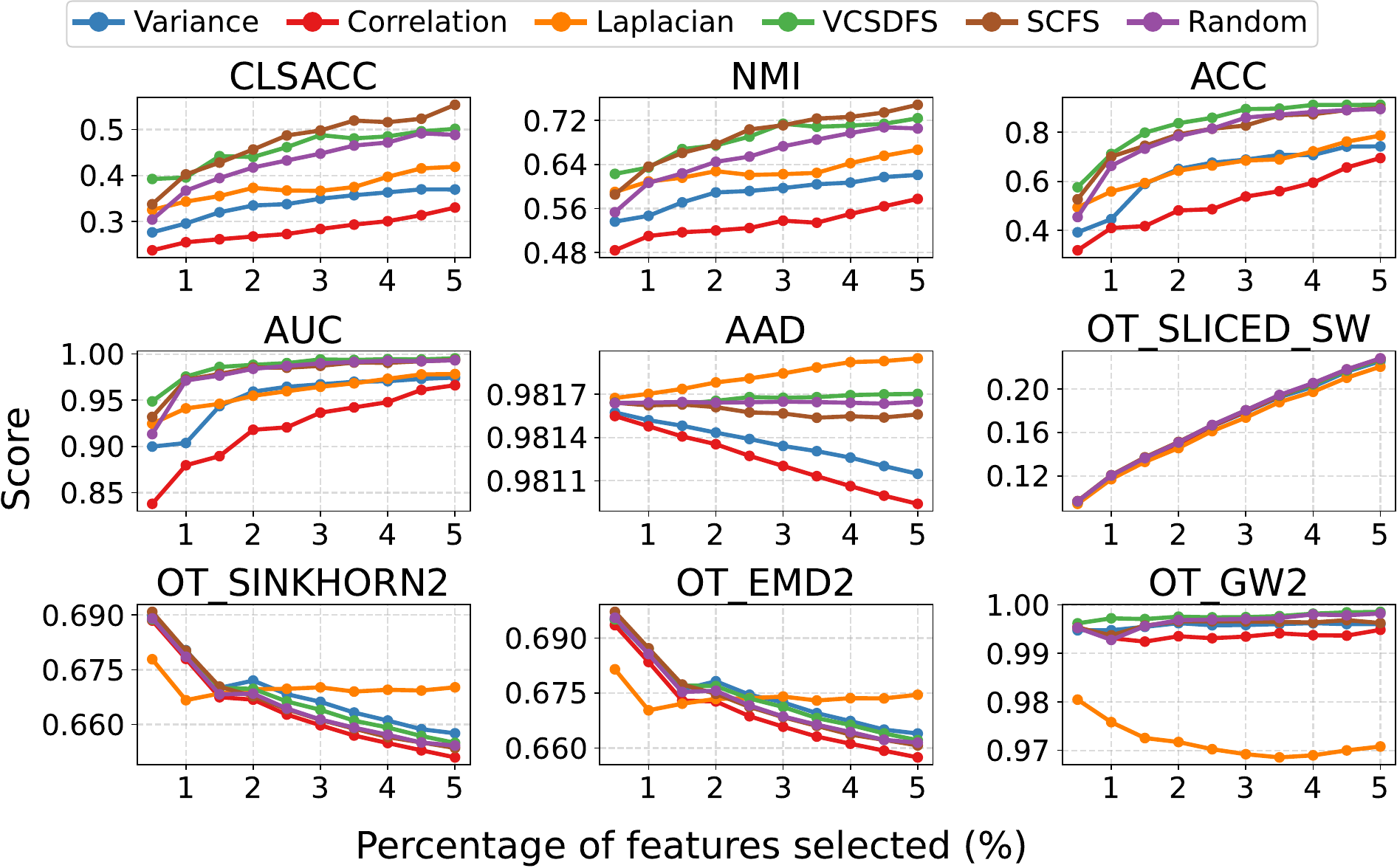}
    \caption{Performance curves of the nine evaluation metrics on \textit{ORL} as a function of the percentage of features selected.}
    \label{fig:ORL_5Percent}
\end{figure}

\begin{figure}[!htbp]
    \centering
    \includegraphics[width=0.9\linewidth]{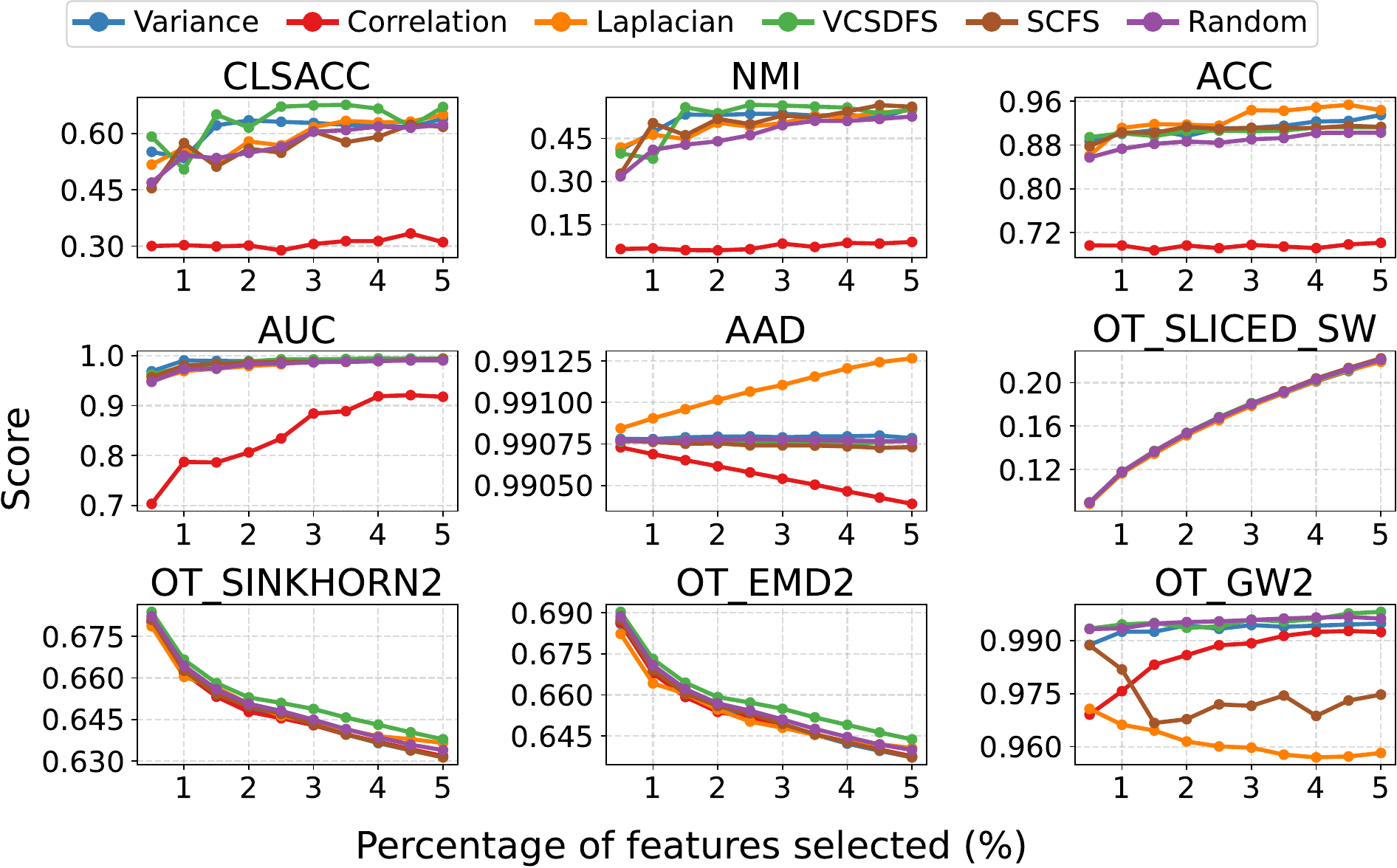}
    \caption{Performance curves of the nine evaluation metrics on \textit{lung} as a function of the percentage of features selected.}
    \label{fig:lung_5Percent}
\end{figure}

\begin{figure}[!htbp]
    \centering
    \includegraphics[width=0.9\linewidth]{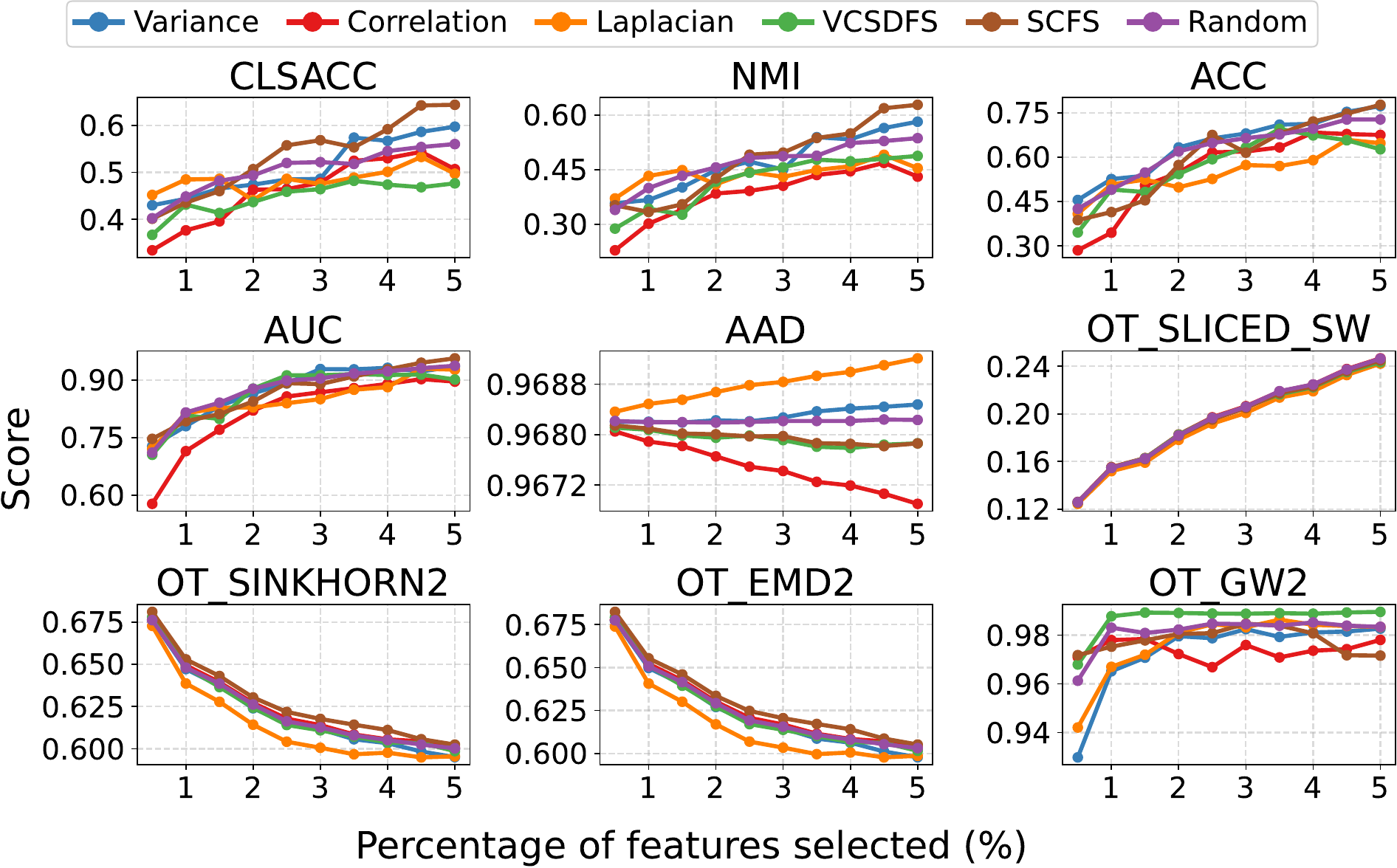}
    \caption{Performance curves of the nine evaluation metrics on \textit{lung\_discrete} as a function of the percentage of features selected.}
    \label{fig:lung_discrete_5Percent}
\end{figure}

\begin{figure}[!htbp]
    \centering
    \includegraphics[width=0.9\linewidth]{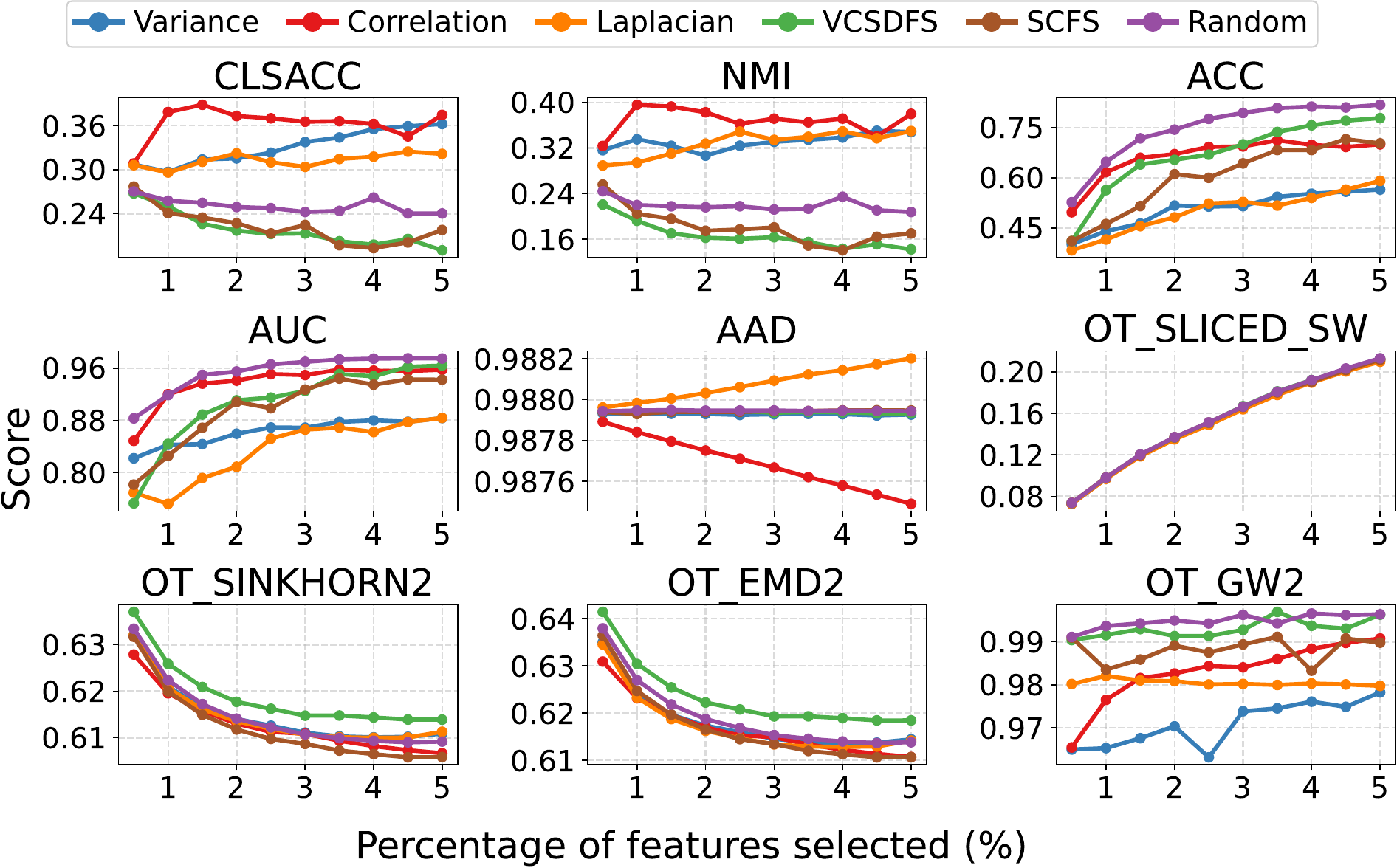}
    \caption{Performance curves of the nine evaluation metrics on \textit{warpAR10P} as a function of the percentage of features selected.}
    \label{fig:warpAR10P_5Percent}
\end{figure}

\begin{figure}[!htbp]
    \centering
    \includegraphics[width=0.9\linewidth]{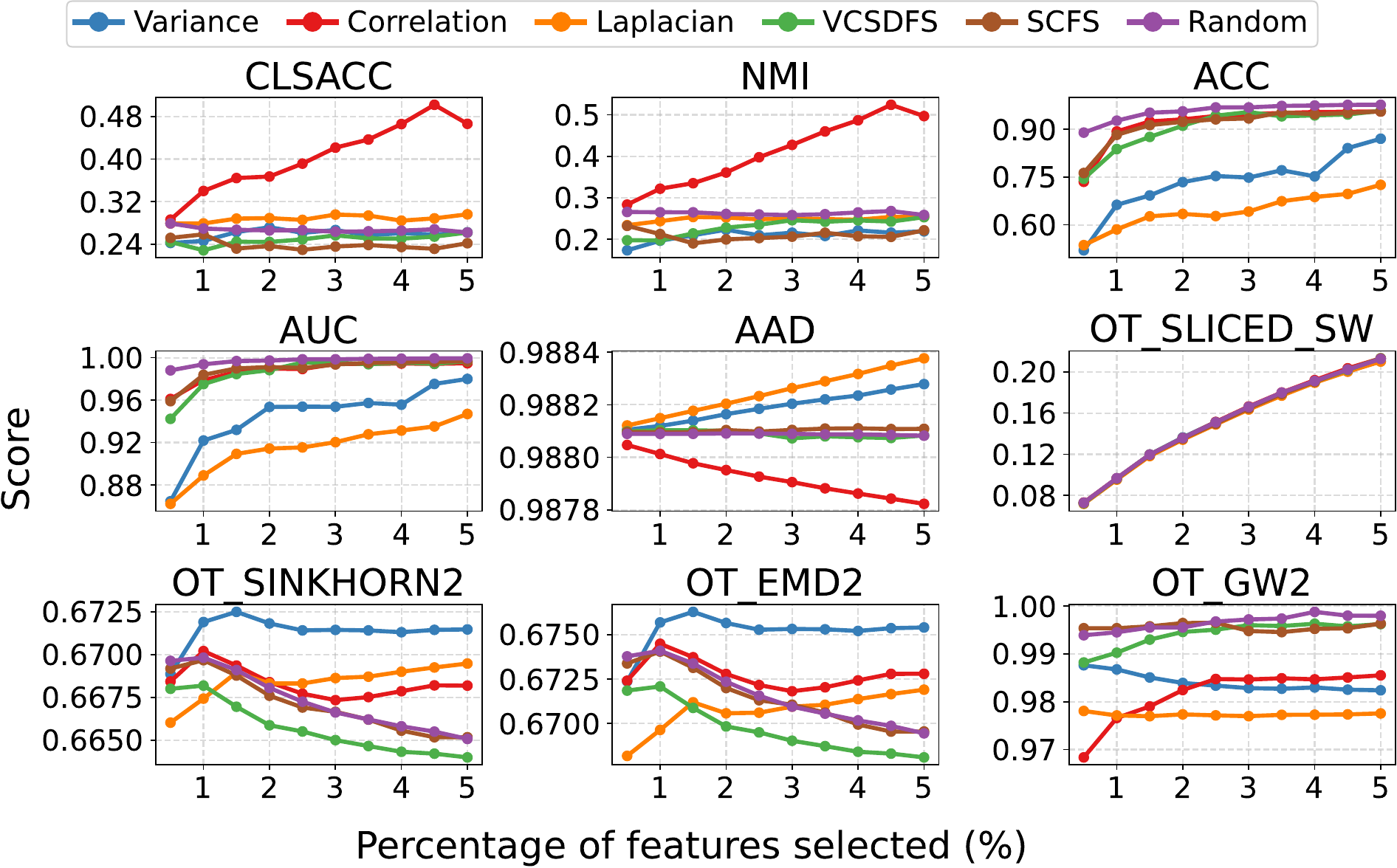}
    \caption{Performance curves of the nine evaluation metrics on \textit{warpPIE10P} as a function of the percentage of features selected.}
    \label{fig:warpPIE10P_5Percent}
\end{figure}

\begin{figure}[!htbp]
    \centering
    \includegraphics[width=0.9\linewidth]{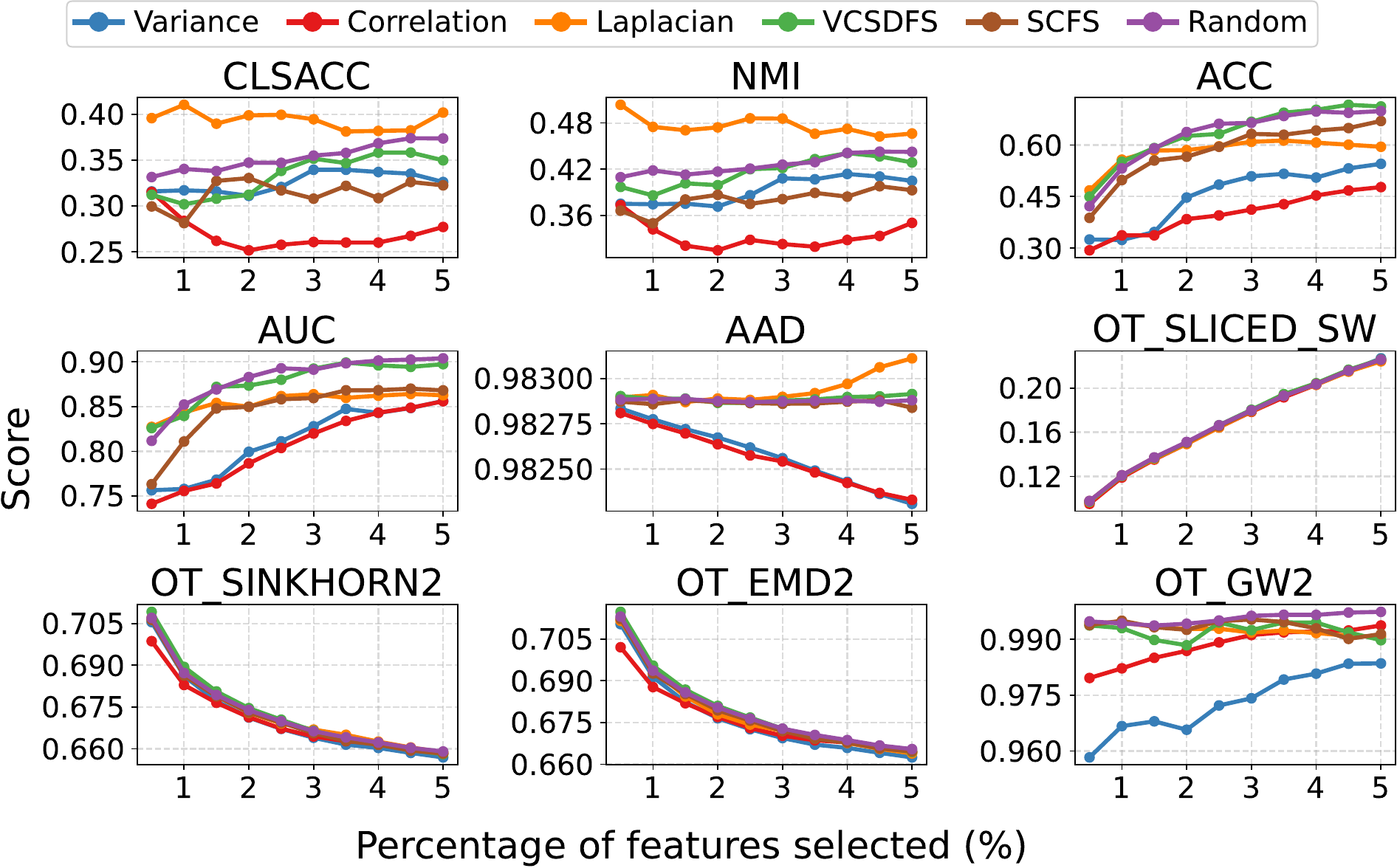}
    \caption{Performance curves of the nine evaluation metrics on \textit{Yale} as a function of the percentage of features selected.}
    \label{fig:Yale_5Percent}
\end{figure}

Figures~\ref{fig:COIL20_5Percent}--\ref{fig:Yale_5Percent} complement Fig.~\ref{fig:Isolet_5Percent} with the performance curves of the nine evaluation metrics on the remaining benchmark datasets. Layout and conventions follow Fig.~\ref{fig:Isolet_5Percent}: each panel shows one evaluation metric as a function of the percentage of features selected, higher is better for all measures, and the optimal transport distances are converted to similarities.

\clearpage
\subsection{Rankings Produced by OT\_SLICED\_SW} \label{app:sliced}
\begin{figure}[!htbp]
    \centering
    \includegraphics[width=0.7\linewidth]{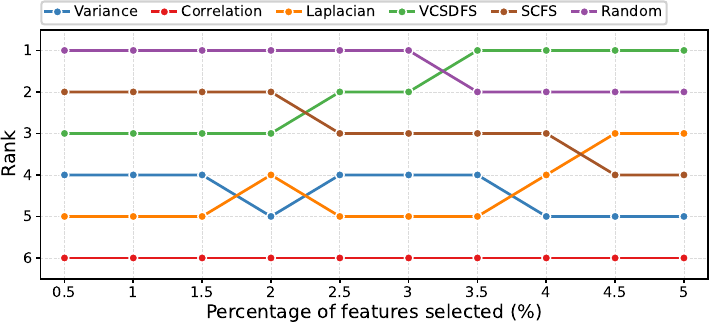}
    \caption{Ranking produced by \textit{OT\_SLICED\_SW} on \textit{COIL20} as a function of the percentage of features selected.}
    \label{fig:COIL20_Sliced}
\end{figure}

\begin{figure}[!htbp]
    \centering
    \includegraphics[width=0.7\linewidth]{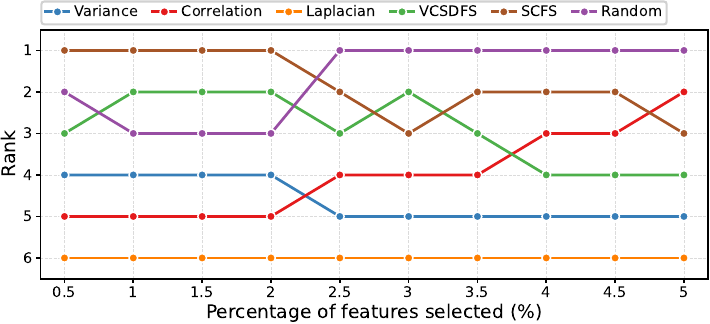}
    \caption{Ranking produced by \textit{OT\_SLICED\_SW} on \textit{ORL} as a function of the percentage of features selected.}
    \label{fig:ORL_Sliced}
\end{figure}

\begin{figure}[!htbp]
    \centering
    \includegraphics[width=0.7\linewidth]{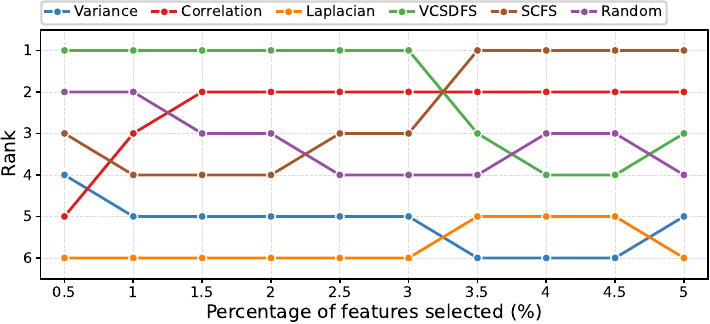}
    \caption{Ranking produced by \textit{OT\_SLICED\_SW} on \textit{lung} as a function of the percentage of features selected.}
    \label{fig:lung_Sliced}
\end{figure}

\begin{figure}[!htbp]
    \centering
    \includegraphics[width=0.7\linewidth]{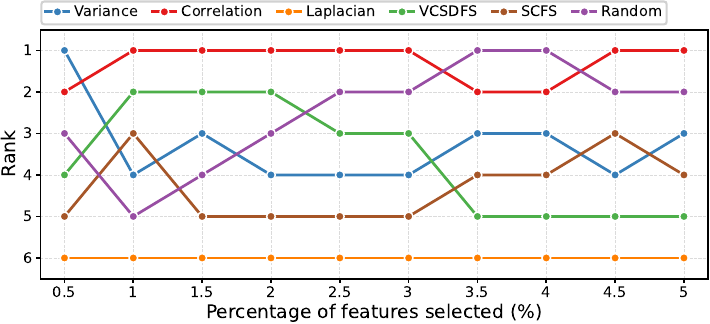}
    \caption{Ranking produced by \textit{OT\_SLICED\_SW} on \textit{lung\_discrete} as a function of the percentage of features selected.}
    \label{fig:lung_discrete_Sliced}
\end{figure}

\begin{figure}[!htbp]
    \centering
    \includegraphics[width=0.7\linewidth]{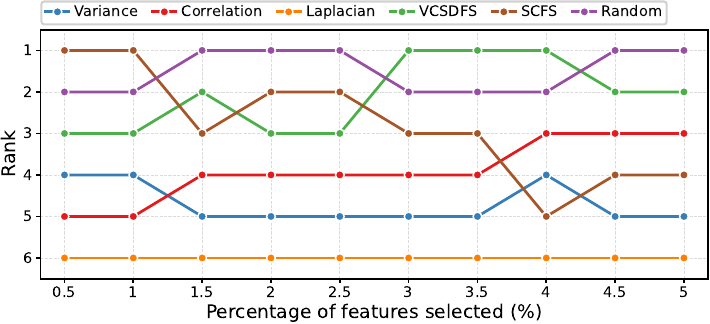}
    \caption{Ranking produced by \textit{OT\_SLICED\_SW} on \textit{warpAR10P} as a function of the percentage of features selected.}
    \label{fig:warpAR10P_Sliced}
\end{figure}

\begin{figure}[!htbp]
    \centering
    \includegraphics[width=0.7\linewidth]{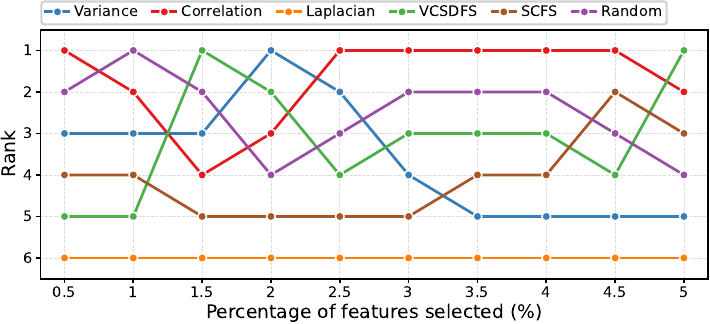}
    \caption{Ranking produced by \textit{OT\_SLICED\_SW} on \textit{warpPIE10P} as a function of the percentage of features selected.}
    \label{fig:warpPIE10P_Sliced}
\end{figure}

\begin{figure}[!htbp]
    \centering
    \includegraphics[width=0.7\linewidth]{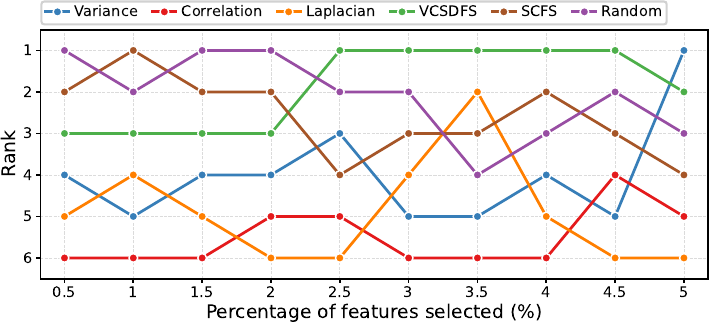}
    \caption{Ranking produced by \textit{OT\_SLICED\_SW} on \textit{Yale} as a function of the percentage of features selected.}
    \label{fig:Yale_Sliced}
\end{figure}

As the \textit{OT\_SLICED\_SW} curves in Figures~\ref{fig:COIL20_5Percent}--\ref{fig:Yale_5Percent} are too compact to distinguish the six feature selection methods, Figures~\ref{fig:COIL20_Sliced}--\ref{fig:Yale_Sliced} report the rankings induced by \textit{OT\_SLICED\_SW} on the remaining benchmark datasets, following Fig.~\ref{fig:Isolet_Sliced}. In all plots, $1$ represents the highest rank and $6$ represents the lowest rank.

\end{document}